\documentclass[twoside,11pt]{article}

\usepackage{jmlr2e}
\usepackage{enumitem}
\usepackage{amsmath}

\newcommand{\indep}{\rotatebox[origin=c]{90}{$\models$}}

\usepackage{listings}
\usepackage{xcolor}

\definecolor{codegreen}{rgb}{0,0.6,0}
\definecolor{codegray}{rgb}{0.5,0.5,0.5}
\definecolor{codepurple}{rgb}{0.58,0,0.82}
\definecolor{backcolour}{rgb}{0.95,0.95,0.92}

\lstdefinestyle{mystyle}{
    backgroundcolor=\color{backcolour},   
    commentstyle=\color{codegreen},
    keywordstyle=\color{magenta},
    numberstyle=\tiny\color{codegray},
    stringstyle=\color{codepurple},
    basicstyle=\ttfamily\footnotesize,
    breakatwhitespace=false,         
    breaklines=true,                 
    captionpos=b,                    
    keepspaces=true,                 
    numbers=left,                    
    numbersep=5pt,                  
    showspaces=false,                
    showstringspaces=false,
    showtabs=false,                  
    tabsize=2
}
\ShortHeadings{Trying to Outrun Causality in Machine Learning (preprint)}{Vowels, M.J.}
\firstpageno{1}

\begin{document}

\title{Trying to Outrun Causality in Machine Learning: Limitations of Model Explainability Techniques for Identifying Predictive Variables}

\author{\name Matthew J. Vowels \email matthew.vowels@unil.ch \\
}

\editor{}

\maketitle

\begin{abstract}
Machine Learning explainability techniques have been proposed as a means of `explaining' or interrogating a model in order to understand why a particular decision or prediction has been made. Such an ability is especially important at a time when machine learning is being used to automate decision processes which concern sensitive factors and legal outcomes. Indeed, it is even a requirement according to EU law. Furthermore, researchers concerned with imposing overly restrictive functional form (\textit{e.g.}, as would be the case in a linear regression) may be motivated to use machine learning algorithms in conjunction with explainability techniques, as part of exploratory research, with the goal of identifying important variables which are associated with an outcome of interest. For example, epidemiologists might be interested in identifying `risk factors' - \textit{i.e.} factors which affect recovery from disease - by using random forests and assessing variable relevance using importance measures. However, and as we demonstrate, machine learning algorithms are not as flexible as they might seem, and are instead incredibly sensitive to the underling causal structure in the data. The consequences of this are that predictors which are, in fact, critical to a causal system and highly correlated with the outcome, may nonetheless be deemed by explainability techniques to be unrelated/unimportant/unpredictive of the outcome. Rather than this being a limitation of explainability techniques \textit{per se}, we show that it is rather a consequence of the mathematical implications of regression, and the interaction of these implications with the associated conditional independencies of the underlying causal structure. We provide some alternative recommendations for researchers wanting to explore the data for important variables.

\end{abstract}


\section{Introduction}
 Many, if not most, machine learning (ML) algorithms are difficult to interrogate, and this difficulty has earned them a reputation for being `black boxes' \citep{Rudin2019}. This is demonstrably problematic, particularly because of the arbitrary nature of the functions being learned. It can result in bias against minority groups, and algorithmic decisions which are heavily influenced by culturally sensitive or legally protected characteristics such as race, age, gender, or sex \citep{Hardt2016, locatello2019fairness, cao2019, liu2019, howard2018, rose2010, gendershades}. In response to concern surrounding bias and opacity of ML driven decision processes, research into model explainability techniques has thrived. Such techniques aim to identify which input dimensions are being used to make which decisions, so that humans can diagnose problematic decisions, or alternatively justify why a certain decision has been made \citep{Lundberg2017, Lundberg2020, Sani2020, Chen2020b, Han2020, Aas2019, Strumbelj2014}. In cases where such a decision concerns the evaluation of some personal attributes and for which there exist legal consequences (for example, a credit application), the EU states that `The data subject should have the right... to obtain an explanation of the decision reached...' \citep{EU2016}, making some form of model explainability essential.

It is, however, important to understand the limitations of machine learning algorithms and explainability techniques.  Generally, model explainability techniques are `advertised' as a means to interrogate the model only, \textit{not} as a means to reliably infer something about the world, and we would agree with this. A random forest, for example, may provide a convenient means to automate various decision making processes across a range of applications. We would call such applications as `practical' - they are not intended to be used to understand real-world phenomena, they are only intended to fulfil a function to reduce cost or improve the efficiency of a decision making process or application. Explainability techniques can be used in this situation to glean which variables are being used by the algorithm to make its decisions.

Given that one might expect an ML algorithm to leverage \textit{any} and \textit{all} correlations available to satisfy a particular learning goal (\textit{e.g.}, minimizing the mean squared error function), one might be forgiven for also expecting that explainability methods provide us with a means to identify variables which are strongly statistically correlated/associated with the outcome. If such an interpretation were possible, it might, for instance, be useful at the initial exploratory phase of a research project intended to develop a computational model of a certain phenomenon. Indeed, before we can build a causally realistic model of the phenomenon, it is important to be able to identify and account for all potentially causally (or indeed, predictively) relevant variables first. Even though we well know that `correlation is not causation', correlated variables which are not directly causally related to an outcome of interest may nonetheless play a critical role in the wider system. We refer to such an application of random forests as `scientific', because they would be being used to derive understanding about some phenomenon of interest.

There are many examples of researchers across a range of domains using explainability techniques `scientifically', \textit{i.e.}, to infer something about the strength of associations in the real-world in order to further a scientific understanding of the real-world. In particular, random forests with explainability methods such as random forest importances, or Shapley values \citep{Lundberg2020} have seen application in the domains of psychology \citep{Vowels2021JSM, Vowelsinfidelity, Joel2020}\footnote{In the interests of full disclosure - the current author has previously used model explainability for this purpose.}, genetics \citep{Goldstein2011}, epidemiology \citep{Orlenko2021, Khalilia2011}, drug-discovery \citep{Luna2020}, and many others \citep{Strobl2007}. Indeed, Yarkoni and Westfall \citep{Yarkoni2017} discuss ways to interpret machine learning models in empirical settings, and argue that they may be used to `help gain a deeper understanding of the general structure of one's data.'

Unfortunately, and as we will show, machine learning models and model explainability techniques cannot be used reliably to infer correlations, and the purview of the explainability should be restricted to be completely \textit{local to the model}. For example, even if a variable importance measure deems a certain variable to be unimportant, the variable itself may actually be highly correlated with the outcome (and therefore predictively or causally relevant/important). Indeed, in our experiments we create a dataset containing variables between which there exist important relationships (both correlational and causal). We show that machine learning techniques `miss' these key variables, at least insofar as the explanations deem them to be unimportant. This is consequence of the interaction between otherwise flexible machine learning algorithms and the underlying structure in the data. As such, we cannot use explainability techniques in conjunction with ML algorithms to reliably inform us even of correlations (let-alone causality). The consequences of this are potentially severe: As a result of the interaction between causal structure and our ML algorithms, we might infer from our explainability output that key risk variables in epidemiological contexts are unimportant, or that potential intervention variables are unassociated with an outcome of interest. These problems notwithstanding, our goal is not to discount machine learning altogether. On the contrary: Machine learning models can be used both in application (\textit{e.g.}, to automate a decision process), or be integrated into a causal model with known structure in order to estimate effects without the imposition of assumptions regarding functional form \citep{Vowels2021}. However, if one wishes to use machine learning algorithms in the latter sense, some strong inductive biases are required to ensure the problems described in this work do not apply (\textit{i.e.}, we must be confident in the causal structure we specify).

In this paper we adopt a causal perspective to help us understand and explain the limitations of machine learning with model explainability techniques. In particular, we examine the sensitivity of these techniques to the underlying structure of the Data Generating Process (DGP). Although the results derive from interactions between structure and regression in general (\textit{i.e.}, independent of the particular regression algorithm used), we empirically demonstrate the almost inescapable dependence that linear models (and their coefficients), random forests (both variable importance measures and Shapley values), and neural networks / MultiLayer Perceptrons (and the associated Shapley values) have on the underlying structure, highlighting how highly correlated variables may nonetheless be deemed unimportant, even when using powerful ML algorithms and state-of-the-art explainability techniques. Whilst the sensitivity of the coefficients of linear models can be reasonably understood given knowledge of the underlying structure \citep{Vowels2021}, the results for the importance measures and Shapley values (with both random forests and MultiLayer Perceptrons) have, to the best of our knowledge, not been described before. This work is therefore important in highlighting how the output of explainability techniques can almost seem arbitrary - they have some discernibly regular dependence on the underlying structure, but even knowing the structure makes it difficult to predict the relative predictor importances. Our conclusion is that it is not possible to `outrun causality in machine learning', and that it always important to understand the possible interactions between the algorithm and the underlying causal structure in the data. Of course, we may rarely have access to the true structure, making the early stages of research challenging. If we wish to scour the data to identify variables which are relevant to a particular phenomenon of interest, what options are we left with? We propose two options: bivariate mutual information, and causal discovery techniques, and we introduce these techniques and discuss how they can be used. 

The paper is structured as follows:

\begin{itemize}[leftmargin=.5in]
    \item Section \ref{sec:background}, Background: We review some relevant background theory relating to causality, Directed Acyclic Graphs, \textit{d}-separation, regression, random forests, multilayer perceptrons, and explainability.
    \item Section \ref{sec:methodology}, Methodology: We provide details on the experiments, including the datasets, algorithms, and explainability techniques.
    \item Section \ref{sec:experiments}\footnote{Complete code for the experiments is openly available at \url{https://github.com/matthewvowels1/ML_structural_interactions}}, Results and Discussion: We demonstrate how, for a given causal graph, both random forest importances and Shapley value techniques for machine learning algorithms cannot be used to reliably infer anything about the presence of correlations/associations in the data. The chosen experimental setting is a 9 variable regression task.
    \item Section \ref{sec:recommendations}, Recommendations: We provide two ways to develop an understanding of a phenomenon at the exploratory stages of a research project, specifically, using bivariate mutual information, and causal discovery).
    \item Section \ref{sec:conclusion}, Conclusion: We summarise the work and provide some suggestions for future work.
\end{itemize}

\section{Background}
\label{sec:background}
\subsection{Graphs}
In order to explore the inability of machine learning models and explainability techniques to reliably inform us of correlations in the data, we take a causal perspective. In particular, we use Directed Acyclic Graphs (DAGs) to operationalize the dependence of machine learning models on the underlying structure. In this section we introduce the relevant concepts for the subsequent exploration. Much of the content is adapted from \citet{Vowels2021DAGs}. Interested readers are encouraged to consult other resources on graphical models and causal inference such as \cite{Hunermund2021, Peters2017, Koller2009}.

\begin{figure}[t!]
\centering
\includegraphics[width=1\linewidth]{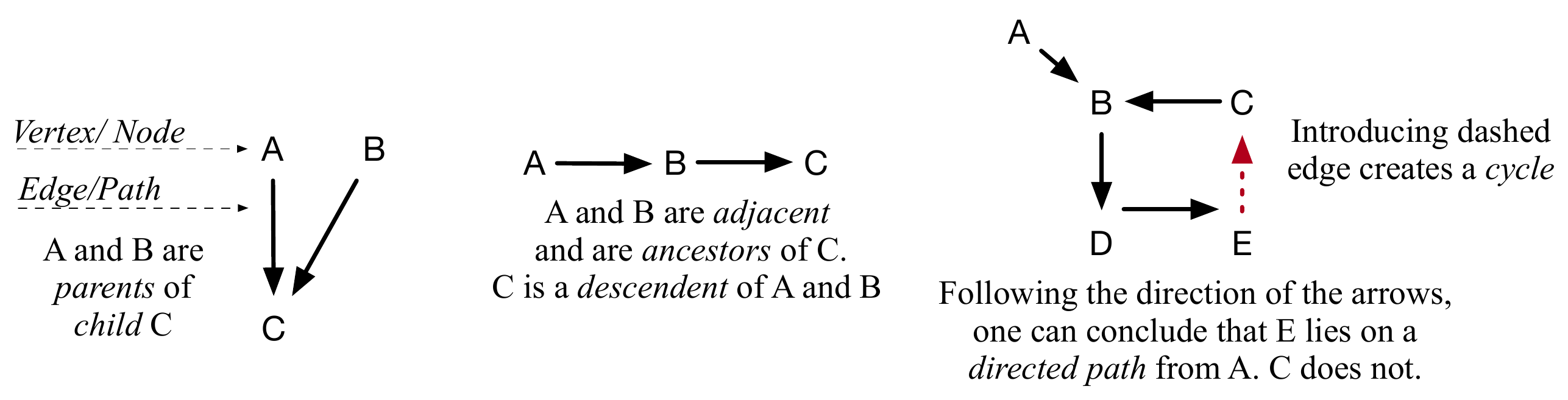}
\caption{A visual representation of some of the notation.}
\label{fig:defs}
\end{figure}

 We follow a similar formalism to Peters et al. \citep{Peters2017} and Strobl \citep{Strobl2018}. These definitions are quite dense, and we therefore provide a figure illustrating them in Fig.~\ref{fig:defs}. Many  may already be familiar with Structural Equation Models (SEMs), which represent a popular subclass of Structural Causal Models (SCMs), whereby the SEMs are usually constrained to encode linear dependencies. SCMs enable an expression-based (rather than graphical) representation of structure. For example, the chain structure represented as the graph $X_i \rightarrow X_j \rightarrow X_k$ (which tells us that $X_i$ causes $X_k$ via a mediator $X_j$) is a graphical abstraction of the following system of equations in an SCM:

\begin{equation}
    \begin{split}
        X_i := f_i(U_i),\\
        X_j := f_j(X_i, U_j),\\
        X_k := f_k(X_j, U_K).
    \end{split}
    \label{eq:scm}
\end{equation}

In the case of SEM, the functions $f$ are linear. Furthermore, the use of the assignment operator `$:=$' makes explicit the asymmetric nature of these equations. In other words, they are not to be rearranged to solve for their inputs. Also of note are the $U_{\{i,j,k\}}$ terms, which represent unobserved exogenous noise variables which are usually omitted from the graphical representation. Including them would involve additional causes for each of the $X$ nodes \textit{e.g.}, $U_i \rightarrow X_i$.

 Working with the graphical portrayals of structural relationships provides an intuitive and immediately visually comprehensible representation. As such, we devote some time to defining the relevant notation and terminology. A directed graph $\mathcal{G}(\mathbf{X}, \mathcal{E})$ represents a joint distribution $P_{\mathbf{X}}$ as a factorization of $d$ variables $\mathbf{X} = \{X_1, ..., X_d\}$ using $d$ corresponding \textit{nodes/vertices} $v \in \mathbf{V}$ and connecting, directed edges $(i, j) \in \mathcal{E}$, where $(i, j)$ indicates a directed edge between $v_i$ and $v_j$. If two vertices $i$ and $j$ are connected by an edge we call them \textit{adjacent}, and, can also denote this in terms of the corresponding variables $\mathbf{X}$ as $X_i \rightarrow X_j$ or $X_i \leftarrow X_j$. If all edges are directed, and there are no cycles, we have the  class of \textit{Directed Acyclic Graphs} (DAGs). 

We can define a \textit{parent} $pa_j$ as a vertex $v_i$ with \textit{child} $v_j$ connected by a directed edge $X_i \rightarrow X_j$ such that $(i,j) \in \mathcal{E}$ but $(j,i) \notin \mathcal{E}$. Further upstream parents are \textit{ancestors} of downstream \textit{descendants} if there exists a directed path constituting $i_{k} \rightarrow j_{k+1}$ for all $k$ in a sequence of vertices. An \textit{immorality} or \textit{v-structure} describes when two non-adjacent vertices are parents of a common child. A \textit{collider} is a vertex where incoming directed arrows converge.

DAGs are assumed to fulfil the Markov property, such that the implied joint distribution factorizes according to the following recursive decomposition, characteristic of Bayesian networks \citep{Pearl2009}:

\begin{equation}
    P(\mathbf{X}) = \prod_i^d P(X_i |pa_i) .
    \label{eq:decomposition}
\end{equation}

Eq.~\ref{eq:decomposition} tells us that the joint probability of the system can be calculated as the product of the probabilities of each of the $d$ variables, conditional on its parents. Taking the $\log$ of this expression makes the quantity computatable as a sum, instead of a product. By way of example, the likelihood of the graph on the left hand side of Fig.~\ref{fig:defs} can be computed as: $P(A,B,C) = P(A)P(B)P(C|A,B)$, and the log-likelihood can be computed as: $\log P(A,B,C) = \log P(A) + \log P(B) + \log P(C|A,B)$. 
The decomposition relates to the notion of $d$-separation. Two vertices $X_i$ and $X_k$ are $d$-separated by the set of vertices $\mathbf{S}$ if $X_j \in \mathbf{S}$ in any of the following structural scenarios \citep{Peters2017}:

\begin{equation}
    \begin{split}
        X_i \rightarrow X_j \rightarrow X_k \; \; \; \mbox{(chain)}\\
        X_i \leftarrow X_j \leftarrow X_k  \; \; \; \mbox{(chain)}\\
        X_i \leftarrow X_j \rightarrow X_k  \; \; \; \mbox{(fork)}\\
    \end{split}
    \label{eq:dsep1}
\end{equation}

They are also $d$-separated if neither $X_j$ \textit{nor any of the descendants of} $X_j$ are in set $\mathbf{S}$ in the following structural scenario:

\begin{equation}
    \begin{split}
        X_i \rightarrow X_j \leftarrow X_k \; \; \; \mbox{(collider)}
    \end{split}
    \label{eq:dsep2}
\end{equation}

If the DAG's $d$-separation properties hold (an assumption of faithfulness - see below), they imply Markovian conditional independencies in the joint distribution, which can be denoted as $X_i \indep_{P_{\mathbf{X}}} X_k | X_j$. In words, node $X_i$ is statistically independent of $X_k$ given $X_j$. In terms of the DAG, disjoint (\textit{i.e.}, non-overlapping) sets of variables $\mathbf{A}$ and $\mathbf{B}$ are $d$-separated by disjoint set of variables $\mathbf{S}$ in graph $\mathcal{G}$ if $\mathbf{A} \indep_{d-sep} \mathbf{B} | \mathbf{S}$ \citep{Peters2017}, and are, conversely $d$-connected if this conditional independence in the graph does not hold. 

To transform these relationships from graphical/mathematical relationships to \textit{causal} relations, the \textit{Causal} Markov Condition is imposed, which simply assumes that the arrows represent causal dependencies and that there are no unobserved or unmodelled confounders \citep[p.105-6]{Peters2017}. It is then common to use the DAG framework as a means to represent domain knowledge relating to the underlying DGP. The ultimate benefit of the graphical and structural model frameworks is that they, at least in principle and under some strong assumptions, enable us to use observational data to answer scientific questions such as `how?', `why?', and `what if?' \citep{Pearl2018TBOW}. If a domain expert has a theory about the structure underlying a given phenomenon, they may represent this theory graphically using a DAG. 

In the presence of statistically dependent unobserved variables, the graph is said to be \textit{semi}-Markovian, because some of the implied graphical conditional independencies may not hold in practice as a result of the additional dependencies induced by the unobserved variables. These unobserved variables are usually denoted with $U$, as in \ref{eq:scm}, but if the Markov condition holds they are usually omitted from the graph for convenience (their presence does not affect the entailing machinery of the graph). However, when the graph is semi-Markovian, it is actually more convenient to indicate these relationships graphically than in a system of equations, by including the dependence between $U$ vertices graphically. For example,  a curved, dashed, bidirectional link between the observed variables can be used to indicate the presence of an unobserved confounder $U$.

\subsection{Regression}
As statistics or machine learning practitioners we may be confronted with a dataset $\mathcal{D} = \{y^i, \mathbf{x}^i\}_{i=1}^{N}$ sampled from $\mathcal{P}= \{Y,\mathbf{X}\}$. In words, we have a dataset with $N$ observations of some outcome variable of interest $Y$, and a set of predictors $\mathbf{X}$. In terms of notation, we use \textbf{bold} to denote multi-dimensionality, lower-case to denote (\textit{e.g.}) realisation $y^i$ of upper-case random-variable Y, $\mathcal{P}$ to denote a population distribution, and $N$ the sample size. This notation is compatible with the notation used above for graphs, such that $x_j^i$ is specific realisation $i$ of variable $X_j$, which in turn may constitute a node in graph $\mathcal{G}$.

In regression, we might be concerned with estimating the conditional expectation of $Y$ using the set of variables $X$ with regression model having parameters $\theta$. Specifically, we often wish to estimate:

\begin{equation}
    \mathbb{E}[Y|\mathbf{X}=\mathbf{x}] = \int y f(y|\mathbf{x}) dy,
    \label{eq:regression}
\end{equation}
where $f$ is the conditional density. Importantly, note that this expectation involves a conditioning statement $Y|\mathbf{X}$. If $Y$ were binary, the expectation would be equivalent to $P(Y|\mathbf{X}=\mathbf{X}$. This has important implications for \textit{d}-separation as we described in the previous section because conditioning on variables can induce independence or dependence (depending on the structure). We can use a regression model (\textit{e.g.}, a random forest) $m$ to approximate $\mathbb{E}[Y|\mathbf{X}=\mathbf{x}] \approx m(\mathbf{x})$ using an empirical sample or subset of our dataset $\mathcal{D}$. Of course, with parametric assumptions we can fit a linear function $m$ via Ordinary Least Squares. However, we assume readers are already familiar with the construction of such as model (following the usual $Y = \beta_0 + \beta_1X_1 + ... +\beta_KX_K + \epsilon$ form where $K$ is the number of variables in the set $\mathbf{X}$ and $\epsilon$ is exogenous noise, or model error). It is, however, useful to review some theory behind much more flexible models, because it is in light of their flexibility that the results in the experiment section are so unintuitive. We therefore provide a brief review of both the random forest, and the MultiLayer Perceptron (MLP), below.

\subsubsection{Random Forests}

Let us begin begin with a basic decision tree, the model for which can be represented as \citep[pp.546]{murphy2}:

\begin{equation}
    m(\mathbf{x}) = \sum_{m=1}^M w_m\mathbb{I}(\mathbf{x} \in R_m) = \sum_{m=1}^M w_m \phi(\mathbf{x};\mathbf{v}_m)
\end{equation}

Here, $w_m$ is the mean response of the $m$th region $R_m$, which represents a partition of the input space. $\mathbf{v}_m$ denotes the variable being used to split on, and the associated threshold $t$ for this split. For example, $x_2 \leq t_2$ indicates a split using variable $x_2$ if it is less than or equal to threshold $t_2$). One can interpret this model as  `an adaptive basis-function model', where the basis functions contain this variable and threshold information, and are given by $\phi(\mathbf{x};\mathbf{v}_m)$. The weights then tell us the outcome value in each of the $M$ regions. The model can be trained to identify the best variable and thresholds to split on according to a variation of the squared error\citep[pp.548]{murphy2} $\sum_i^N(y_i - \bar{y})^2$ where here, $\bar{y}$ represents the average of the outcome variable for the associated set of data. As such, the random forest `carves' up the input space according to these splits, thereby deriving a highly non-linear mappying from the predictors to the outcome. 

The popular extension of the decision tree, namely the random forest, reduces the variance of estimates from a decision tree by averaging together the predictions from many similar trees. By training $K$ different trees using different subsamples of our dataset (subsamples of both datapoints and variables), we can derive a powerful algorithm for prediction.

\subsubsection{MultiLayer Perceptrons (MLPs)}
The MLP \citep{haykin2} is a type of neural network, comprising multiple linear layers and non-linear activations. Assuming a set of $K$ predictors $\mathbf{x}$, the output of the first two layers $\mathbf{l}_1$ and $\mathbf{l}_2$ can be represented as \citep{goodfellow}:

\begin{equation}
    \begin{split}
        \underset{(n \times H_1)}{\mathbf{l}_1}  = \sigma(\underset{(n \times K)}{\mathbf{x}}\underset{(K \times H_1)}{\mathbf{W}_1^T} + \underset{(H_1)}{\mathbf{b}_1}),\\
        \underset{(n \times H_2)}{\mathbf{l}_2}  = \sigma(\underset{(n \times H_1)}{\mathbf{l}_1}\underset{(H_1 \times H_2)}{\mathbf{W}_2^T} + \underset{(H_2)}{\mathbf{b}_1}),\\
    \end{split}
\end{equation}

Here, $\mathbf{x}\mathbf{W}_1^T$ is a matrix multiplication between a matrix of $n$ datapoints for $K$ predictors $\mathbf{x}$ and a matrix of learnable weights $\mathbf{W}_1$, $\mathbf{b_1}$ is a bias or offset term, and $\sigma$ is a non-linearity (\textit{e.g.}, a $tanh$, sigmoid, or ReLU()$=max(0, x)$ function). It can be seen from the construction that the linear function serves not only as an affine transformation (scale, translation, rotation, \textit{etc.}), but also as a means to change the dimensionality of the representation. The final layer $l_L$ can be designed such that its `shape' is $(n\times 1)$, \textit{i.e.}, the same dimensionality as the sample of values for the outcome variable $Y$. In words, we take a set of predictors and recursively process them according to a set of these non-linear layers, until they can be mapped to the desired outcome. One may stack any number of these layers (we use two layers in the experiments), one after the other, and have the inner layers operating at a much higher dimensionality than the dimensionality of the set of predictors we started with (we use a dimensionality of 100 in the experiments). MLPs thus facilitate the learning of highly complex, non-linear functions, and have been shown to be `universal function approximators' \citep{hornik2}, which, loosely, means that they can approximate any function.

The MLP is generally trained according to gradient descent, where the weight matrices $\mathbf{W}$ and the bias vectors $b$ are updated according to their impact on a specified loss function $\mathcal{L}$  (\textit{e.g.} mean squared error):

\begin{equation}
    \begin{split}
        \mathcal{L} = n^{-1} \sum_i^n ||\sigma(\sigma(\mathbf{x}^i\mathbf{W}_1^T) \mathbf{W}_2^T) - y^i||^2
    \end{split}
\end{equation}

Using relatively elementary calculus (chain rule) one can compute the extent to which each weight or bias parameter and update them according to a learning rate. Interested readers are directed to the accessible overview of deep learning by Goodfellow, Bengio, and Courville \citep{goodfellow}.

\subsection{Model Explainability}

\subsubsection{Random Forest Importances for Model Explainability}
One of the most common ways to derive feature importances for random forests is via the use of impurity measures. In this work, we consider continuous outcome variables (\textit{i.e.}, we use regressors, rather than classifiers), and in this case the measure of impurity is usually the mean squared error (others are possible, \textit{e.g.}, the mean absolute error). For each of the decision trees in the random forest, the importances can be calculated based on the decrease in impurity (\textit{i.e.}, the improvement in performance) for each node, weighted by the probability of using that node in a particular tree, and these improvements can be averaged across data samples \citep{breiman2001, Strobl2007}. 

In the empirical literature, it is common to use these importances as proxies for variable importance \textit{outside} the model. For example \citep{Joel2020} used machine learning importance measures to infer the presence of associations between variables pertinent to relationship quality in the real-world. The logic seems to be that if a random forest finds a variable useful in making a prediction, then there exists some (potentially non-linear) association in the real world. Indeed, given the bootstrapped nature of random forests, and the way that they can carve up the input space to define highly non-linear mappings between predictors and outcome, we can understand this thought process. Unfortunately, in spite of their flexibility, they are nonetheless constrained according to the structure in the data, and variables which appear unimportant, may actually be important and highly statistically associated.

\subsubsection{Shapley Values for Model Explainability}
The Shapley value explainability methods derive from the seminal game theoretic work of Lloyd Shapley \citep{Shapley1953}. The methods conceive of a regression task as a collaborative game, where each of the predictor variables represents a player. The goal of the game is to maximise the regression performance (or, equivalently, to minimize the regression error), and the explanation quantifies the degree to which each player (\textit{i.e.}, predictor) contributes to this goal. Of course, the role that each predictor plays is difficult to directly ascertain, because it is `collaborating' with other predictors at the same time (in the form of multiplicative interactions $X_1 \cdot X_2$, for example) in specific and complex ways which are largely determined by the ML algorithm itself. The Shapley value methods therefore disentangle these complex contributions by evaluating the impact that each possible combination of predictors has on the model output. The result is a per-predictor, per-datapoint estimation of the impact on model performance, thus providing a fine-grained summary of model behaviour. Interested readers are directed to the recent papers by Lundberg and colleagues \citep{Lundberg2017, Lundberg2019, Lundberg2020}. It is this level of fine-grained information which gives Shapley values a distinct advantage over random forest importances. Furthermore, they have also been shown to be more reliable, in general, than random forest importances \citep{Lundberg2020}, and can be derived for almost arbitrary model classes. Indeed, in the experiment section we will use the Shapley value techniques to derive estimates of variable importances from neural networks.

\begin{figure}[t!]
\centering
\includegraphics[width=0.65\linewidth]{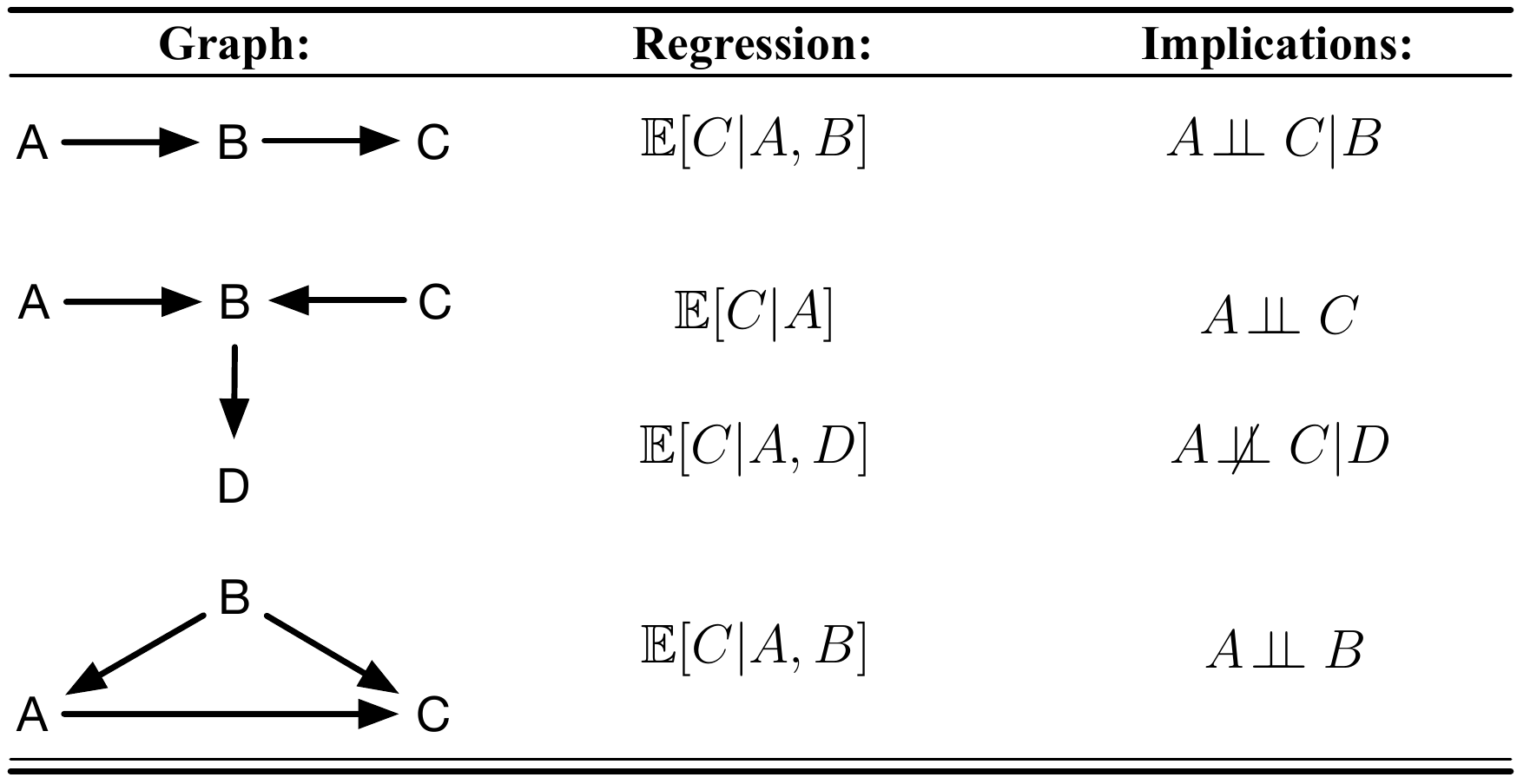}
\caption{In the first graph (a chain structure), all the variables are statistically dependent. However, if we are interested in predicting $C$ from $A$ and $B$, we `block' the path from $A$ to $C$ when we condition on $B$, inducing independence. This would result in $A$ being unimportant for the regression $\mathbb{E}[C|A,B]$. In the middle graph, $A$ is already independent of $C$ because of the collider structure at $B$. The regression $\mathbb{E}[C|A]$ does nothing to affect this independence structure. However, by conditioning on $D$, we are conditioning on a descendent of a collider, which renders $A$ important for the regression $\mathbb{E}[C|A,D]$. Finally, while the regression $\mathbb{E}[C|A,B]$ renders $A$ independent of $B$ (which is useful for `blocking the backdoor path from $A$ to $C$ via $B$'), the conditions do not prevent $B$ or $A$ from being important to the regressor.}
\label{fig:dsep}
\end{figure}

\subsection{Regression and Structure - Possible Explanations}
In order to glean an understanding for why the results reported in later sections are seemingly so structurally dependent requires an understanding of the conditional independencies implied by the underlying graph.Before we present the results, we therefore devote some time to discuss a couple of simple examples. In addition, Fig.~\ref{fig:dsep} provides some additional instances for consideration.  First, let us look more closely at the important implications of the \textit{d}-separation rules.

Starting with a simple dataset $\mathcal{D} \sim \mathcal{P} = \{Y, X_k\}_{k=1}^{K}$ where $k$ denotes the variable index, consider the graphs ($K=2$) given by:

\begin{equation}
    \begin{split}
        X_1 \xrightarrow{\beta_1} Y \xleftarrow{\beta_2} X_2  \\
    X_1 \xrightarrow{\beta_1} X_2 \xrightarrow{\beta_2} Y 
    \end{split}
    \label{eq:examples}
\end{equation}

As with our later experiments, let us assume that the dependencies are linear, to keep things simple. We have indicated with the path coefficients $\beta$ the true (population level) strengths of the dependencies between variables in these graphs. Let us also assume that the true values of these path coefficients are all equal to one, \textit{i.e.}, $\beta_1=\beta_2 =1$ for both graphs.

In the context of regression, we may wish to estimate $\mathbb{E}[Y|\mathbf{X}]$. This expectation is, itself, dependent on a model of the conditional distribution of $Y|\mathbf{X}$ (as per Eq.~\ref{eq:regression}). For both graphs above, our regression implies a conditional density $Y|X_1,X_2$, and this has different implications for each of the two graphs above, and the implications can be understood via the \textit{d}-separation rules. 

For the first graph in Eq.~\ref{eq:examples}, the regression entailing the conditional density for $Y|X_1,X_2$ does nothing to interfere with the conditional independencies encoded by the original graph. Namely, $X_1 \indep X_2$ regardless of whether we are conditioning on the $X$ variables as part of the regression or not. Indeed, if we were to undertake a linear regression, or fit the data using a random forest, we would expect the associated coefficients/importances to be approximately equal, reflecting the fact that the true dependencies between these variables ($\beta_1 = \beta_2 = 1$) are also equal. In this case, the coefficients from a linear regression will be unbiased estimates of the true path coefficients, \textit{i.e.}, $\hat{\beta}_1 \approx \beta_1$, and $\hat{\beta}_2 \approx \beta_2$.

On the other hand, in the second graph, the conditioning in our regression results in $X_2 \indep Y | X_1$. In words, even though there is a clear dependency structure between $X_1$ and $Y$, by conditioning on $X_2$ this structure is `broken', and $X_1$ and $Y$ are rendered independent. Imagine the true dependencies are linear and that $\beta=1$. This means the effect of both $X_1$ and $X_2$ on $Y$ is also one (according to the multiplication of the path coefficients). Instead, the coefficients of a linear regression $Y|X_1, X_2$ will estimate $\hat \beta_1 \approx 0 \not\approx \beta_1$ which is incorrect, and will correctly estimate $\hat \beta_2 \approx 1 \not\approx \beta_2$.

\section{Methodology}
\label{sec:methodology}

Having reviewed some relevant background material, we take a moment to discuss the motivation for the following experiments.\footnote{Note that full code for the experiments is provided at \url{https://github.com/matthewvowels1/ML_structural_interactions}} On the face of it, given the nature of random forests and neural networks, there is little to concern us that a random forest would be necessarily prevented from leveraging any or all useful correlations between the predictors $\mathbf{X}$ and the outcome $Y$. Indeed, it seems like the opposite might be more likely: The fact that a random forest can arbitrarily partition the input space according to multiple bootstrapped decision trees, where the details of the partitioning are driven by a very general cost function (such as the squared error) perhaps encourages us to think that the algorithm can do whatever it wants to leverage any and all statistical associations in the data. Similarly, the fact that the neural network is a universal function approximator, and can expand the dimensionality of the predictors arbitrarily might lead us to believe that it has relatively free-reign or equal opportunity to use any and all useful variables. In turn, then, we might also expect explainability techniques (such as random forest importance measures, or Shapley values) to yield predictor importance levels which are relatively agnostic to the structure of the DGP which led to the observations with which the models were trained.

However, and as we will see in the Section \label{sec:experiments}, in spite of the flexibility of random forests and neural networks, the methods are nonetheless sensitive to the interaction between the conditioning statements in the associated regression being undertaken, and the underlying structure of the DGP. As a result, the use of variable importance measures (including Shapley value techniques) does not help us to reliably identify predictively useful variables.

\subsection{Data}

\begin{figure}[t!]
\centering
\includegraphics[width=0.7\linewidth]{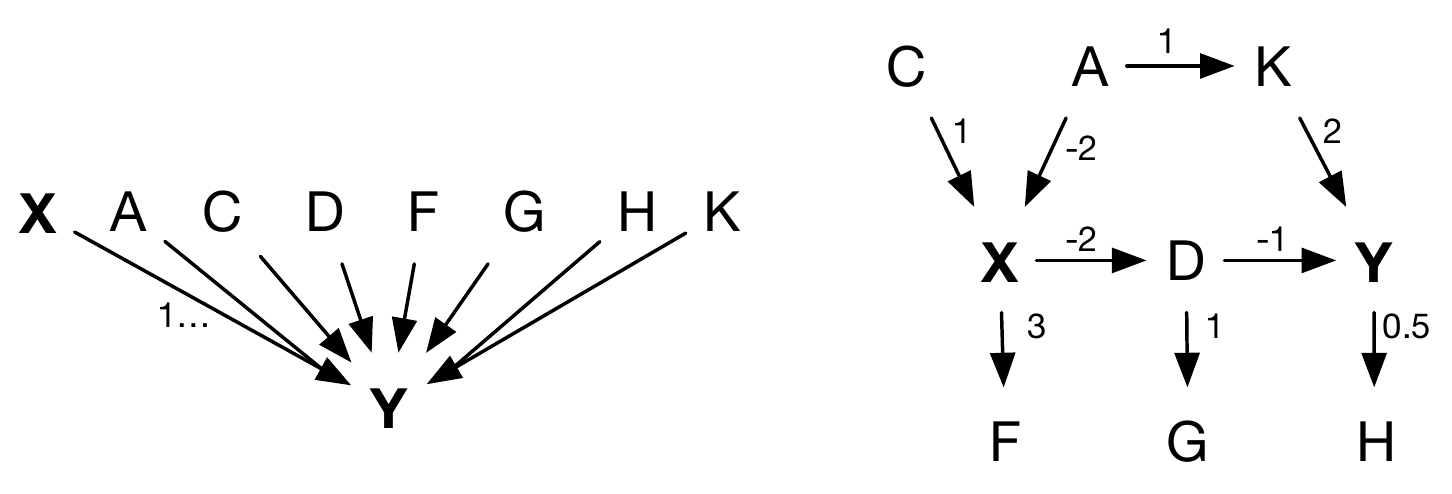}
\caption{The causal structures of the two datasets used in the experiments. On the left, the graph is trivial - all predictors are independent of each other and cause the outcome $Y$. The path coefficients are all one. On the right, the structure of the second dataset is more complex. In a slight abuse on notation, in both graphs variables $X$ and $Y$ are highlighted in \textbf{bold}. This to emphasize that in practice, we might be particularly interested in the influence a particular predictor has on the outcome, when that predictor is just one in a system of many. In the left graph, the influence of $X$ on $Y$ is the same as all the others (and equal to one), whereas in the right graph, the influence of $X$ on $Y$ is equal to $-2 \times -1 = 2$ according to the multiplication of the path coefficients for the mediated path $X\rightarrow D \rightarrow Y$. }
\label{fig:graphs}
\end{figure}

We create two datasets, each with nine variables (eight predictors and one outcome variable). The structures of these datasets are shown in Fig.~\ref{fig:graphs}. Both datasets are generated according to linear functional relationships, and the corresponding path coefficients are denoted in the Figure. In a slight abuse of notation, variables $X$ and $Y$ are highlighted in \textbf{bold} in these graphs, because we assume them to be variables of interest for the sake of the experiments. For instance, $X$ might refer to some kind of `risk' variable, which we expect, as domain experts, to affect outcome $Y$. We use a sample size $N=10,000$ to avoid estimation variability due to sample size. 

It can be seen the the first dataset (Fig.~\ref{fig:graphs}, left) has a trivial, exogenous error structure, with independent predictors. The second dataset (Fig.~\ref{fig:graphs}, right) is based on one from Peters et al. \citep{Peters2017} and Vowels \citep{Vowels2021}, and is more complex, containing:

\begin{itemize}
    \item Direct effects, \textit{e.g.}, $D \rightarrow Y$.
    \item Mediated effects, \textit{e.g.},  $X \rightarrow D \rightarrow Y$).
    \item `Backdoor' paths \citep{Pearl2009}, \textit{e.g.}, $X\leftarrow A \rightarrow K \rightarrow Y$, where $X$ is linked to $Y$ via an indirect, non-causal path.
    \item Colliders, \textit{e.g.}, $C \rightarrow X \leftarrow A$.
\end{itemize}  

The system of equations (the SCM) representing this second dataset is:

\begin{equation}
    \begin{split}
        C \sim \mathcal{N}(0,1),& \; \; \; \; \;       A \sim \mathcal{N}(0, 0.8),\\
        U_K \sim \mathcal{N}(0, 0.1),& \; \; \; \; \; K = A + U_k, \\
        U_X \sim \mathcal{N}(0, 0.2),& \; \; \; \; \;  X = C - 2A + U_X,\\
        U_F \sim \mathcal{N}(0, 0.8),& \; \; \; \; \;  F = 3X + U_F,\\
        U_D \sim \mathcal{N}(0, 0.5),& \; \; \; \; \; D = -2X+U_D, \\
        U_G \sim \mathcal{N}(0, 0.5),& \; \; \; \; \; G = D + U_G,\\
        U_Y \sim \mathcal{N}(0,0.2),& \; \; \; \;  \; Y = 2K - D + U_Y,\\
        U_H \sim \mathcal{N}(0, 0.1),& \; \; \; \; \; H = 0.5Y + U_H.
    \end{split}
\end{equation}

Here, $\sim \mathcal{N}(\mu, \sigma)$ denotes that observations for these variables are samples from a normal distribution with mean $\mu$ and standard deviation $\sigma$. Despite the increased complexity of this graph, in our opinion it is not so complex to be implausibly representative of real-world causal structures. The dataset is split 60/40 into train and test proportions. Given that previous work has highlighted the sensitivity of random forest importance measures to the variance of the data, we standardized all data before use \citep{Strobl2007}. This also makes comparison between different explainability measures more comparable, particularly as the absolute values of the bivariate correlations are, by their definition, constrained to fall between 0 and 1. The bivariate correlations are shown in Table~\ref{tab:pearsoncorrs}. It can be seen that all variables are highly (and statistically significantly) correlated with the outcome. This is intentional - they are all important variables, and we wish to understand whether machine learning methods can help us identify them.

\begin{table}[h!]
\centering
\footnotesize
  \caption{Bivariate Pearson correlations and $p$-values, $R(p)$, for the right-hand DAG in Figure \ref{fig:graphs}.}
  \label{tab:pearsoncorrs}
  \begin{tabular}{lrrrrrrrr}         \hline
 $r(p)$ &         $X$ &       $D$ &      $A$ &       $K$ &      $C$ &     $F$ &       $G$ &      $H$ \\ \hline
  $Y$ & .92(.00) & -.94(.00)& -.60(.00) & -.59(.00) & .76(.00) & .91(.00) & -.93(.00) & 1.00(.00) \\
  \hline
  \end{tabular}
\end{table}

\subsection{Models / Algorithms}
We provide results for bivariate correlations, Linear Regression (\textbf{LR}), Random Forest (\textbf{RF}), and MultiLayer Perceptron (\textbf{NN} - for Neural Network). It is generally known that the default parameters of random forests perform well across a range of applications, without the need for hyperparameter tuning \citep{Probst2019}, and as such, we use the default settings in the scikit-learn package \citep{sklearn}. Similarly, rather than undertaking an exhaustive hyperparameter search for the MLP, we stay close to the default parameters and verify that the test performance is comparable to that of the random forest. The dimensionality of the hidden layers is set to be 100, the number of layers set to 2 (with one additional outcome layer), the activation is chosen to be ReLU, we use the Adam \citep{adam} optimizer with an adaptive learning rate starting at $1\times10^{-3}$, and trained for 200 iterations. 

\subsection{Explainability Techniques}

We provide the bivariate correlations between each of the predictors and the outcome, denoted `\textbf{bi-corrs}' in the results. For the linear regression we simply provide the coefficient values as measures of predictor importances, these are denoted `\textbf{LR-coefs}' in the results. Indeed, linear regression is straightforward to interpret in this regard. For the random forest we provide both importances derived according to the built-in node impurity method in scikit-learn - denoted `\textbf{RF-imps}' in the results - as well as Shapley values using the SHAP (SHapley Additive exPlanations) `Tree Explainer' package \citep{Lundberg2020} - denoted `\textbf{RF-Shap}' in the results. For the MLP we use the SHAP `Kernel Explainer' package, denoted `\textbf{NN-Shap}' in the results. For both the Tree Explainer and the Kernel Explainer we use a train and test size of 1000 datapoints.

\subsection{Trials and Results Presentation}
In order to demonstrate the interaction between the structure of the second (structurally mode complex) dataset and the models, we undertake a number of analyses, each time removing different variables to understand the concomitant impact on the explanations. In each case, we provide a bar plot showing the relative importance of each variable for each method. In order to make the linear regression coefficients, the Shapley values, and random forest importances more visually comparable, we normalize them to have a range of zero to one (the bivariate correlations are left untouched). The Shapley results are derived to be the absolute values of the per-datapoint impact on model output, averaged over the datapoints. Finally, we provide mean squared errors for each of the algorithms / models.

\section{Results and Discussion}
\label{sec:experiments}

In reality, we are unlikely to have access to the true graph structures as provided in Fig.~\ref{fig:graphs}. In practice, we may also be interested in the relationship (associational or causal) between two variables in particular, and we assume these to be $X$ and $Y$, which are highlighted in bold in the graphs. In order to evaluate the sensitivity of explainability measures to the underlying structure, we can use a dataset for which we know \textit{a priori} that there is a strong association between $X$ and $Y$. For the left graph, the causal effect is equal to that of all other variables (one), for the right graph the effect is comparable to the other variables, and equal to $-2 \times -1 = 2$ according to the multiplication of the coefficients on the mediated path $X\rightarrow D \rightarrow Y$. We therefore also know that these variables should ideally be denoted to be of importance by the explainability techniques. Indeed, if the explainability techniques cannot highlight the presence of a strong association (such as that between $X$ and $Y$), we might easily discount otherwise key variables as being unrelated/unimportant.

\begin{figure}[h!]
\centering
\includegraphics[width=0.7\linewidth]{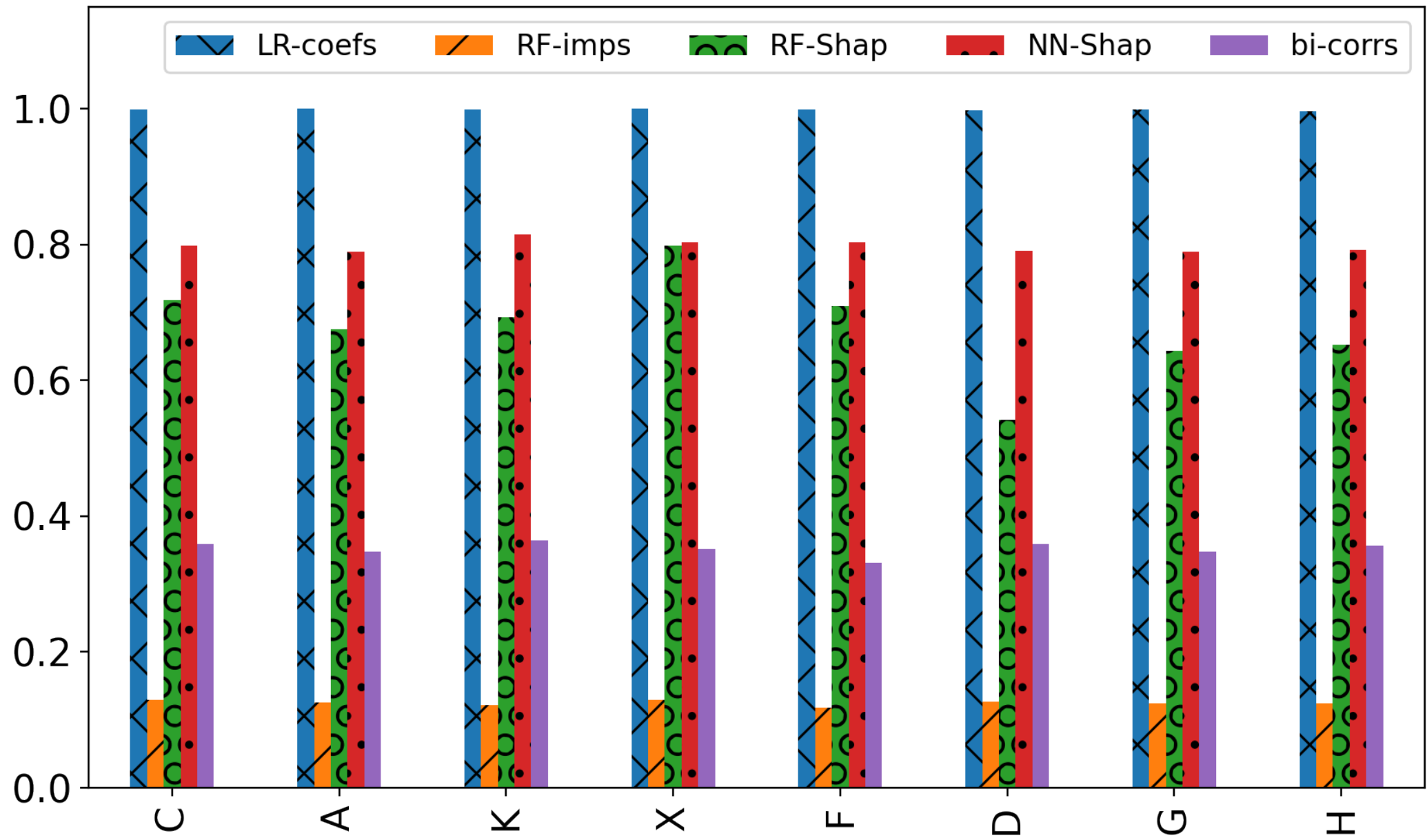}
\caption{Results for linear regression coefficients `LR-coefs', random forest importances `RF-imps', random forest Tree Explainer Shapley values `RF-Shap', neural network Kernel Explainer Shapley values `NN-Shap', and bivariate correlations `bi-corrs'. DAG featured in plot whitespace shows underlying structure. Figure best viewed electronically and in colour. }
\label{fig:plotsimple}
\end{figure}

Let us begin by checking that ML algorithms and explainability techniques are not fooled by trivial/idealistic structures. In Fig.~\ref{fig:plotsimple} we show the results for the simple structure depicted on the left of Fig.~\ref{fig:graphs}. We know from the construction of this dataset that all variables have equal importance, because the causal effect of each variable on the outcome is equal to one. An evaluation of the importances in Fig.~\ref{fig:plotsimple} is reassuring because, indeed, regardless of which method we choose, the importances are rated as equal. even though there are differences in the absolute levels of importance \textit{between} methods, once can nonetheless see that these importances are approximately equal for all variables. To this extent, we have confirmed our expectations that when the structure is trivial (all variables independently causing the outcome), machine learning algorithms can be used to highlight variables of particular importance. There is nothing in the conditional independency structure skewing our assessment of variable importance.

\begin{figure}[h!]
\centering
\includegraphics[width=1.05\linewidth]{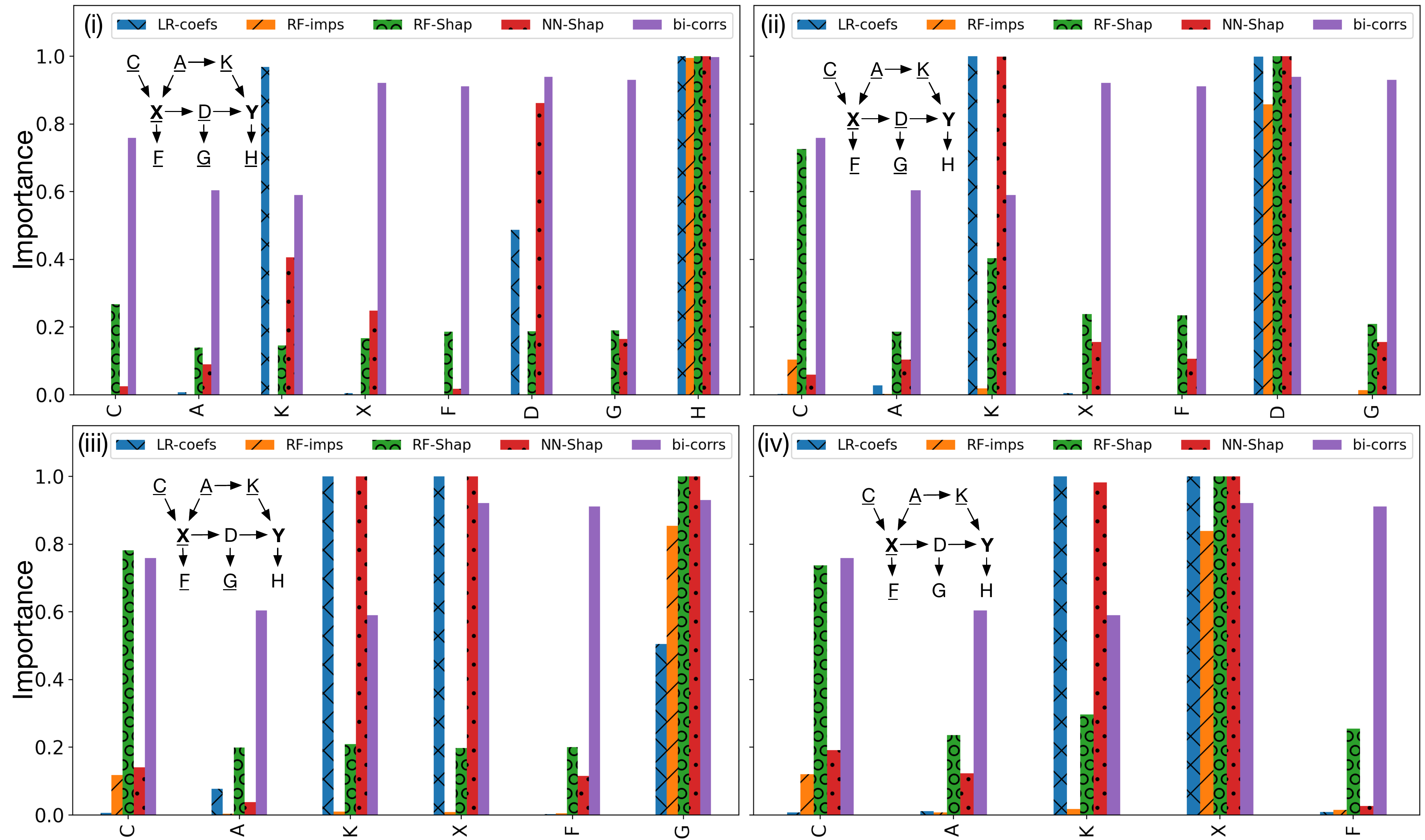}
\caption{Result plots (i-iv) for linear regression coefficients `LR-coefs', random forest importances `RF-imps', random forest Tree Explainer Shapley values `RF-Shap', neural network Kernel Explainer Shapley values `NN-Shap', and bivariate correlations `bi-corrs'. The DAGs featured in the whitespace of each plot denote which variables are included as predictors according to whether the variable is \underline{underlined}. Figure best viewed electronically and in colour. }
\label{fig:plotscomplex}
\end{figure}

The results for the second dataset with more complex structure are shown in Fig.~\ref{fig:plotscomplex}(i-iv). For convenience, these plots feature the DAG in the plot whitespace, with \underline{underlined} variables highlight variables included as predictors in the models. Starting with plot (i), which includes all predictor variables, we see dramatic changes in the relative levels of importance between variables and across methods. For instance, the linear regression coefficients `LR-coefs' on variables $K$ and $H$ are high, followed by $D$, and then all other coefficients are approximately zero. This particular result is easy to explain given knowledge of the true graph - the only paths which have not been blocked by other control variables in the linear model are $H \leftarrow Y$, $D \rightarrow Y$, and $K\rightarrow Y$. At least the linear regression is consistent in this regard, but remember that without this knowledge we would not be able to use the coefficients to infer which variables are important.

Perhaps the next most reasonable set of importances are given by the neural network Shapley values `NN-Shap'. Here, the top three most important variables are, as with the linear regression, $K$, $D$, and $H$. However, $K$'s importance is doubtful. Both the random forest's importances `RF-imps' and Shapley values `RF-Shap' are very misleading - the only variable of note is $H$, with all others having very low importance.

From the first plot alone, we see very strong interaction between the structure and the machine learning explainability results. If we were interested in understanding whether variable $X$ is relevant to $Y$ (which we know it certainly is, because unlike in practice, we have ground truth and simulated the data ourselves) we would have discounted it as unimportant. We could stop there - we have demonstrated that machine learning algorithms, despite their flexibility, are not able to overcoming the constraints deriving from the conditional independencies implied by the underling graph. However, it is of interest to understand how these importances \textit{change}, as variables are removed. In plot (ii) we remove $H$ from the set of predictor variables. Linear regression again provides predictable results - the unblocked paths to the outcome are significant predictors, namely $K\rightarrow Y$ and $D\rightarrow Y$. It would still not be possible to reliably interpret these results without knowledge of the true graph, but they are, at least, consistent with the graph. Again, the neural network provides results which are reasonably consistent with the linear regression, with $K$ and $D$ being highlighted as the most important. Unfortunately, the random forest importances and Shapley values are completely unpredictable: This time, variables $C$ and $D$ are most important. 

In plot (iii) we have removed variables $H$ and $D$. Variable $X$ is now (finally) deemed to be an important variable by the linear regression and the NN. However, this is despite the fact that the descendent of the mediator $G$ is still in the model. This may bias the estimate somewhat, but it is not enough to block the path completely between $X$ and $Y$. As such, the fact that the linear regression coefficients and NN Shapley values indicate importance for $X$ is still reasonable. They also indicate that $G$ is important, which is also reasonable given the open path $G \leftarrow D \rightarrow Y$ without $D$ included as a predictor. Once again, the random forest Shapley value results are somewhat unexplainable, with $C$ being denoted to be the most important variable, followed by $G$. The random forest importances only indicate that $G$ is important.

The final plot (iv) removes $H$, $D$, and $G$. Now we expect unbiased estimates of the causal effect of $X$ on $Y$ by the linear regression, and this is indicated also by the high values for the coefficients on $X$. As before, $K$ is also deemed important, as expected. The NN yields similar results, whereas, once again, the random forest importances and Shapley values are somewhat unpredictable.

\subsection{Summary}

In summary, linear regression, random forests, and neural networks are \textit{all} subject to the effects of underlying conditional independencies associated with structured data. Both linear regression coefficients and neural network Shapley values respected this structure well. Both would, of course, not yield interpretable results without knowledge of the underlying graph, and so both are not recommended if one wishes to identify key/important variables as part of an initial exploration for a research project. But, they are, at least, predictable in their interactions with the structure. In contrast, random forests behaved quite strangely, with the associated importance and Shapley values only vaguely predictable in their interaction with the underlying structure.
 
\section{Recommendations}
\label{sec:recommendations}

Based on the results and discussion presented above, we might be tempted to conclude that machine learning algorithms do not function predictably (in terms of the way they interact with the underlying structure) nor consistently, and therefore constitute an unappealing analytical choice. Indeed, based on our analysis their behaviour was less intuitive than the linear regressor. However, the principal conclusion is that \textit{all} models (including linear regression) are strongly affected by the underlying structure in the data, and so when choosing between approaches we ought to also consider whether machine learning provides other advantages. To this end, the principal advantage of machine learning techniques is that they do not make strong \textit{a priori} assumptions about the functional form relating sets of variables. As a result they often prove to be strong regressors and classifiers, with higher accuracy than more traditional statistical models \citep{Yarkoni2017, Vowels2021}. This advantage should not be understated.

Furthermore, when used \textit{with} causality/structure, it is possible to get the best of the machine learning model, without the unexpected consequences of the interactions between the model and the underlying causal structure. The confluence of machine learning (ML) and causality represents a burgeoning field, as researchers from both domains recognize the potential synergy behind the combination of powerful function approximation techniques, and structural constraints which facilitate fundamental, scientific reasoning \citep{Scholkopf2019, Vowels2021}. Targeted learning approaches \citep{vanderLaan2014} are gaining traction in the epidemiological domains, and these approaches `bake' powerful machine learning models into causal inference tasks in order to get more accurate estimates of causal quantities. Furthermore, other machine learning models aim to infer the causal structure from the data themselves, as part of a latent variable approach to causal inference \citep{Vowels2020b, Zhang2020, Louizos2017b}. Indeed, if we wanted to correctly estimate some of the causal effects in the more complex graph (the task of causal inference), we have a number of additional options. Firstly, we could specify the full graph as an SEM, and estimate it using an SEM solver package. This option has its disadvantages because it requires correct knowledge of the full graph (and SEM methods can be data-hungry to achieve sufficient power). Secondly, we could estimate specific effects of interest using a targeted approach \citep{vanderLaan2014}. This does not need access to the full graph, only a sufficient adjustment set to control for spurious backdoor paths. See \citep{Vowels2021} for a demonstration of the backdoor adjustment \citep{Pearl2009} method with this graph. There are many other causal inference techniques which exist which could also be adapted to this task \citep{Kennedy2020, Yoon2018, Vowels2020b, Wu2021, Zhang2020}. 

These options notwithstanding, it is, in general, difficult to proceed with empirical research in the absence of some prior knowledge about the causal structure. If we wish to substitute machine learning with explainability for a more reliable approach for identifying important predictive and/or causal variables, then we recommend the use of bivariate mutual information, and/or causal discovery approaches and discuss these in turn below.

\subsection{Bivariate Mutual Information}
Of course, if we are interested in which variables are correlated with the outcome, we can simply compute the correlations between each of the variables in the set of considered predictors $\mathbf{X}$, and the outcome $Y$. However, correlation is a linear proxy for statistical dependence (with some slightly unintuitive properties, \citealt{Vowels2021}), and if we wish to identify variables which are associated but which may have some kind of arbitrarily complex, non-linear dependence, we are better off computing the Shannon Mutual Information (M.I.) between the variables. M.I. is a measure of how much information one variable contains about another \citep{cover2006, Kraskov2004, Steeg2012, Steeg2013, Gao2003, Kinney2014, Vowels2021}, and for two variables $X$ and $Y$, is given by:

\begin{equation}
    I(X;Y)=H(X) - H(X|Y).
    \label{eq:mi}
\end{equation}

In Eq.~\ref{eq:mi}, $H(X)$ is the entropy of $X$ and $H(X|Y)$ is the conditional entropy of $X$ given $Y$. Entropy can be interpreted as the uncertainty, surprise, or degree of randomness associated with a variable, and is computed as:

\begin{equation}
-\sum_n^N p(x^n) \log p(x^n)
\end{equation} 

where $x^n$ is the $n$th realisation of variable $X$, and $p(x^n)$ is the probability of observing this value. Actually, the expression above concerns \textit{discrete} variable $X$ (\textit{e.g.}. Bernoulli distributed), and a similar expression exists for continuous variables which is known as the differential entropy. M.I. may also be equivalently defined as:

\begin{equation}
    I(X;Y) = \int p(X,Y) \log \frac{p(X,Y)}{p(X)p(Y)}
    \label{eq:mi2}
\end{equation}

This second expression for mutual information given in Eq.~\ref{eq:mi2} is already written in terms of continuous variables, and it can be seen that it involves the ratio between the joint distribution $p(X,Y)$ and the product of the marginal distributions $p(X)\cdot p(Y)$. If $X\indep Y$ then, by elementary probability $p(X,Y) = p(X) \cdot p(Y)$ (the joint probability is equal to the product of the marginals) and the expression reduces because:
\begin{equation}
\log \frac{p(X,Y)}{p(X)p(Y)} = 0 = I(X;Y) \; \; \; \mbox{when } X \indep Y
\end{equation}

It can be seen that there are no explicit assumptions of parametric form required for the density functions $p$. The mutual information can therefore be estimated analytically assuming \textit{e.g.} that $p(X)$, $p(Y)$, $p(X,Y)$ are jointly Gaussian, or by using some model for these densities (\textit{e.g.}, histograms). Unfortunately, density estimation techniques tend to be quite data-hungry for accurate estimates, and density estimation is, by itself, a challenging task. One non-parametric route which handles mixtures of continuous and discrete/categorical data is based on a nearest-neighbours algorithm \citep{Gao2018, Runge2018}\footnote{A nearest-neighbor-based M.I. independence testing algorithm is implemented in an open-source python package `tigramite' available at \url{https://github.com/jakobrunge/tigramite}.}. The method provides a $p$-value for evaluating the null-hypothesis that the variables are independent, \textit{i.e.} if $p_{val} \leq \alpha$ we infer that variables $X\not\!\perp\!\!\!\perp Y$, where $\alpha$ is the false positive rate.

Once again, the advantage of the M.I. approach is that it does not require the assumption of either parametric distributions, or of linear functional form, and can be used to identify predictors which share information with the outcome of interest.

\subsection{Causal Discovery}

Even though the conditional independency structure of the underlying graph was ultimately responsible for the failure of the machine learning algorithms to identify important variables, we can actually use this structure to our advantage by using algorithms which explicitly test for it. For overviews of causal discovery methods, interested readers are directed to \citep{Vowels2021DAGs, Heinze2018, Glymour2019, Spirtes2016}. 

Consider the graph $A \rightarrow B \leftarrow C$. This graph already implies the conditional independency structure of $A \indep C | \emptyset$ where $\emptyset$ clarifies that we are conditional on the empty set (\textit{i.e.} nothing). This can be ascertained using, \textit{e.g}, the bivariate M.I. measure described in the section above. Similarly, and again by using the M.I. measure, we can ascertain that $A \not\!\perp\!\!\!\perp B$ and that $B \not\!\perp\!\!\!\perp C$.\footnote{Here, $\not\!\perp\!\!\!\perp $ is used to denote dependence.} There is only one graph which is compatible with this knowledge, and that is the true graph $A \rightarrow B \leftarrow C$. We could actually also identify this using a measure of conditional mutual information (which fulfils the role of a conditional independence test), because for this graph $A \not\!\perp\!\!\!\perp C |B$, because $B$ is a collider, and conditioning on $B$ `explains away' the other variables, thereby inducing dependence.

\begin{figure}[h!]
\centering
\includegraphics[width=0.6\linewidth]{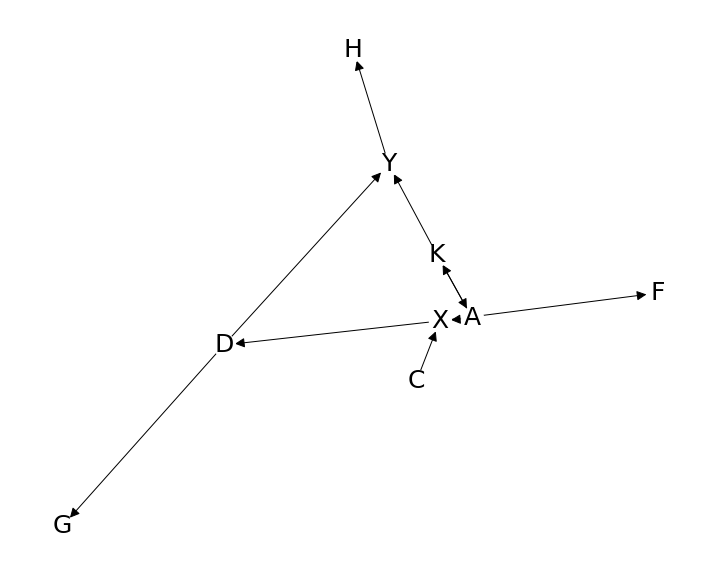}
\caption{Output of the PC algorithm \citep{Spirtes2000}, as implemented in the `pgmpy' python package \citep{pgmpy}. Graph is the CPDAG estimate for data generated according to the right-hand graph in Fig.~\ref{fig:graphs}.}
\label{fig:pcout}
\end{figure}

Similar to the nearest-neighbor-based M.I. pairwise independence test described above, there is also a \textit{conditional} nearest-neighbor-based M.I. independence test \citep{Runge2018}.\footnote{A nearest-neighbor-based conditional M.I. independence testing algorithm is implemented in an open-source python package `tigramite' available at \url{https://github.com/jakobrunge/tigramite}.}. The method provides a $p$-value for evaluating the null-hypothesis that the variables are independent, given a conditioning set, \textit{i.e.} if $p_{val} \leq \alpha$ we infer that $X\not\!\perp\!\!\!\perp Y|\mathbf{S}$, where $\alpha$ is the false positive rate. 

Other graphs are not so straightforward as the example above, and it may not be possible to orient all the edges in the graph. This results in what is known as a Markov Equivalence Class (MEC), which comprises a set of graphs which are indistinguishable from each other using conditional independence tests alone. The conditional M.I. test can therefore be used to test combinations of variables in the set of variables of interest to derive a `CPDAG', which is a Completed Partially Directed Acyclic Graph. In the CPDAG, any edges which cannot be oriented using only conditional independence tests can be left undirected, with all possible directions of undirected edges constituting the MEC. For example taking data generated according to the right-hand graph in Fig.~\ref{fig:graphs}, we can use the well-known PC algorithm \citep{Spirtes2000} in order to derive an estimate for the CPDAG using conditional independencies. We show the corresponding output of the implementation of the PC algorithm openly available in the package `pgmpy' \citep{pgmpy}. Even though the orientation of the graph is different, it can be seen that the graph is almost correct up to the undirected (bidirected in the figure) edges, and the link observed between $A\rightarrow F$ instaed of $X\rightarrow F$. Note that this result was obtained without an exploration of the robustness of the estimated graph to changes in hyperparameters. Specifically, the PC algorithm has a parameter $\alpha$ which is the false-positive rate, and which has a big impact on the discovered edges.\footnote{We have made the code used to derive this estimate openly available at \url{https://github.com/matthewvowels1/ML_structural_interactions}. Note that this example assumes linear functional relationships and Gaussian variable distributions. As such, the conditional independence test used is Pearson correlation.} 

As the number of possible graphs increases super-exponentially with the number of vertices/variables, the number of required conditional independence tests also increases. This makes the search problem extremely computationally expensive, especially if a non-parametric mutual-information type test is used. This has motivated a flurry of other machine learning based approaches to structure learning which are continuously optimized and which are not combinatoric search algorithms. Unfortunately, the majority of these methods are score-based, which means they evaluate each putative graph according to its ability to explain the observations (\textit{e.g}, using variants of the log-likelihood). It is well understood that `fitness to data is an insufficient criterion for validation causal theories' \cite[pp.61]{Pearl2009}, and indeed score based methods have been shown empirically to be sensitive to the scaling of the data (\textit{e.g.}, standardizing the data can have a significant impact on the estimated structure) \citep{Reisach2021, Kaiser2021}.

There are a large number of causal discovery algorithms available, and some of them use more than just conditional independencies and variants of likelihood scores to infer the structure \citep{Vowels2021NSM, Sugihara2012, janzing2009, Kalainathan2020}. This enables them to orient additional edges, thereby providing a set of possible solutions which is smaller than the set of Markov-equivalent graphs derived using only conditional independencies. We recommend a wider exploration of such techniques, particularly during the exploratory stages of research, to identify important variables as well as to provide insight into the structure underlying the observations. As with any data-driven approach, particularly those which implicate causality, we, like Dawid \citep{Dawid2008} and his reference to Bourdieu  warn researchers to beware of "sliding from the model of reality to the reality of the model" \citep{Bourdieu1977}.

\section{Conclusion}
\label{sec:conclusion}

In this work we have seen that flexible, powerful machine learning algorithms are not agnostic to the underlying conditional independency structure of the DGP which yielded the observations. Specifically, linear regression and neural networks were severely affected by the underlying structure, but the associated importances (linear regression coefficients and Shapley values, respectively) were at least consistent with our expectations given knowledge of the true graph. In contrast, random forests were also severely affected by the underlying structure, but in ways which were quite unpredictable given knowledge of the structure. Based on the experiments in this paper, we could recommend neural networks over random forests for identifying important predictors. However, this recommendation is somewhat moot, because even thought neural networks were affected by the underlying structure in ways which were predictable, they nonetheless would not be useful without prior knowledge of said structure. The bottom line is therefore that one cannot `outrun causality in machine learning', and that despite of the powerful function approximation capabilities of machine learning algorithms, they cannot be used to reliably identify important predictive and/or causal variables. The consequences of this can be severe, and can result in missclassifying key predictive or causal variables as unimportant.

Rather than recommending that researchers utilize \textit{predictive approaches} to `help gain a deeper understanding of the general structure of one's data' \citep{Yarkoni2017} (which can, as we have shown, greatly mislead us as to the relevance of certain variables), we provide two opportunities for researchers who are, perhaps, at the early, exploratory stages of a research project, and who are seeking a means to identify important variables. Namely, we recommend mutual information as a means to identify statistical dependence between variables, without the need for assumptions about the  functional form, and without needing to constrain the analysis to parametric distributions. Secondly, we recommend researchers engage with techniques from the domains of causal discovery, in order to provide a means to highlight variables which have statistical relevance and to contextualise such variables within an initial estimate of the causal structure. We would also like to emphasize that rather than practitioners being generally discouraged from using machine learning techniques as a consequence of this work, we instead highlight the potential for machine learning techniques to mitigate the need for unreasonable assumptions about the functional form. Indeed, in some regards it is reassuring that machine learning algorithms, in spite of their `black-box' reputation, are nonetheless constrained according to the rules of regression and the conditional independency structure of the data. Furthermore, if we are able to integrate machine learning algorithms into analyses which adequately account for the underlying structure, we can benefit from the power of the machine learning algorithms without the associated problems demonstrated in this work.

In terms of future work, it would be pertinent to undertake further analyses using a range of other machine learning algorithms, in particular using datasets with alternative structures and categorical outcomes. It would also be of interest to check whether these problems apply to much more complex machine learning models, such as transformers \citep{Vaswani2017} and other large-scale deep learning models (such as those used in natural language processing). In particular, transformers utilize the attention mechanism, which can be interpreted in a similar way to importance measures. It therefore represents an interesting alternative model, which may be subject to the same interference from the conditional independencies in the underlying structure.

\vskip 0.2in
\bibliography{NN}

\end{document}